\documentclass{touchannotator_arxiv}

\usepackage{amsmath,amsfonts,bm}

\def\eqref#1{equation~\ref{#1}}

\def\1{\bm{1}}

\DeclareMathAlphabet{\mathsfit}{\encodingdefault}{\sfdefault}{m}{sl}
\SetMathAlphabet{\mathsfit}{bold}{\encodingdefault}{\sfdefault}{bx}{n}

\usepackage{amsmath,amssymb}
\usepackage{booktabs}
\usepackage{longtable}
\usepackage{graphicx}
\usepackage{multirow}
\usepackage{array}
\usepackage{xcolor}
\usepackage{microtype}
\usepackage{subcaption}
\usepackage{placeins}
\usepackage{needspace}
\usepackage{afterpage}
\usepackage{tikz}
\usetikzlibrary{arrows.meta,positioning,fit,backgrounds}
\usepackage{xr-hyper}
\usepackage{hyperref}
\usepackage{url}

\newcommand{\method}{TouchAnnotator}

\title{TouchAnnotator: Multi-Source Tactile Supervision with Privileged-View Distillation}
\author{%
  Ao Li\textsuperscript{1,2,*}\quad
  Zhihao Yuan\textsuperscript{2,*}\quad
  Chubin Zhang\textsuperscript{1,2}\quad
  Kunrui Huang\textsuperscript{3}\quad
  Sheng Xu\textsuperscript{2,4}\\[3pt]
  Zengjue Chen\textsuperscript{5}\quad
  Junjie Wang\textsuperscript{2,4}\quad
  Jiawei Li\textsuperscript{2}\quad
  Yansong Tang\textsuperscript{1,\ensuremath{\dagger}}\\[8pt]
  {\mdseries\fontsize{10}{12}\selectfont
    \textsuperscript{1}Tsinghua University\quad
    \textsuperscript{2}Joy Future Academy, JD\quad
    \textsuperscript{3}JD Technology\\[2pt]
    \textsuperscript{4}The Chinese University of Hong Kong Shenzhen\quad
    \textsuperscript{5}Jilin University}\\[6pt]
  {\normalfont\fontsize{10}{12}\selectfont
    \textsuperscript{*}Equal contribution\quad
    \textsuperscript{\ensuremath{\dagger}}Corresponding author}%
}
\hypersetup{
  pdftitle={TouchAnnotator: Multi-Source Tactile Supervision with Privileged-View Distillation},
  pdfauthor={Ao Li, Zhihao Yuan, Chubin Zhang, Kunrui Huang, Sheng Xu, Zengjue Chen, Junjie Wang, Jiawei Li, Yansong Tang},
  pdfsubject={Multi-source tactile supervision and privileged-view distillation},
  pdfkeywords={egocentric video, tactile prediction, privileged-view distillation}
}

\newcommand{\arxivresourcelink}[3]{%
  \noindent\raisebox{-0.2em}{#1}%
  \hspace{0.5em}\textsf{\textbf{#2:}}\hspace{0.45em}\url{#3}\par
}
\renewcommand{\arxivresources}{%
  \par\vspace{3mm}%
  {\raggedright\fontsize{9}{13}\selectfont
    \arxivresourcelink{\includegraphics[width=0.025\linewidth]{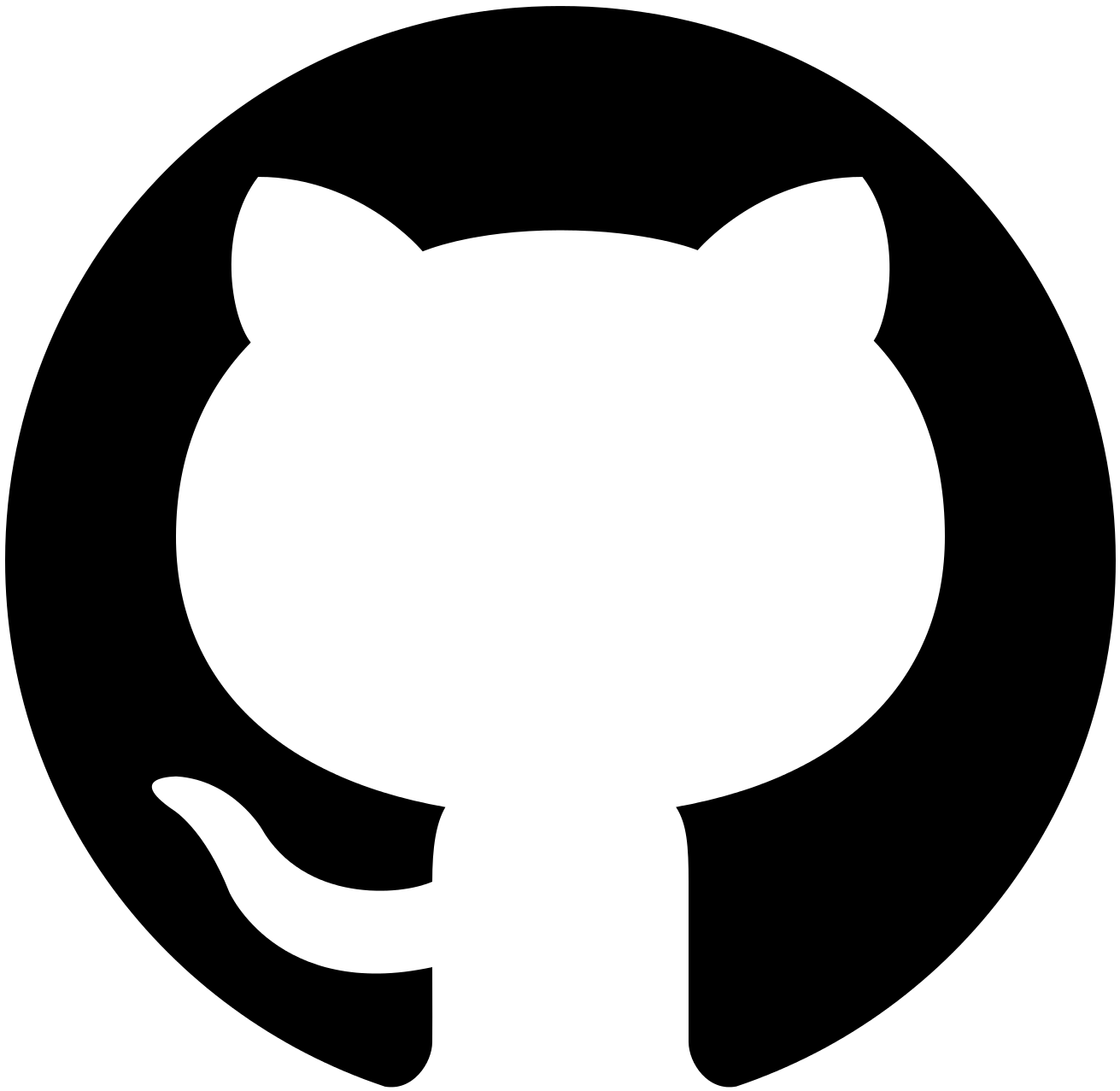}}{GitHub}{https://github.com/RammusLeo/TouchAnnotator}%
    \vspace{3pt}%
    \arxivresourcelink{\includegraphics[width=0.025\linewidth]{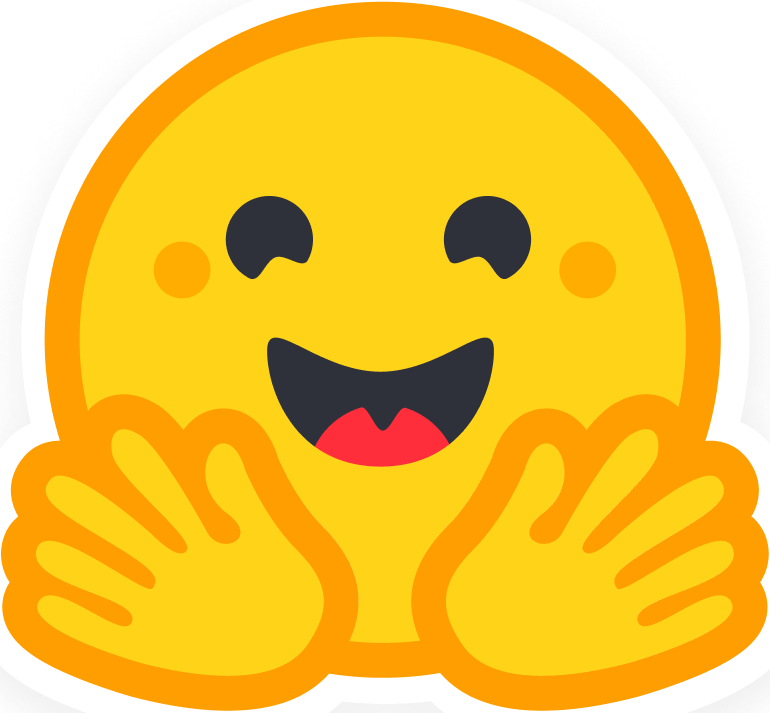}}{HuggingFace}{https://huggingface.co/RammusLeo/TouchAnnotator}%
  }%
}

\begin{document}

\maketitle

\begin{abstract}
Egocentric video captures human--object interaction at scale but rarely includes tactile labels for hidden contact and pressure. We introduce \method{}, an ego-only bimanual tactile annotator trained with multi-source tactile supervision and privileged-view distillation. A canonical hand field unifies glove pressure, geometry-derived human contact, and robot fingertip deformation from six datasets and three embodiments, while source-aware masks preserve valid targets. During training, an EMA teacher observes synchronized ego and wrist views and distills hand-local representations and tactile predictions into a student that uses only ego video and estimated pose at inference. Compact past-clip memory supplies temporal context. We evaluate on EgoTouch, zero-shot FPHA transfer, and 860 synchronized glove recordings. The model transfers to bare-hand contact and produces qualitatively coherent tactile responses across bare-hand, gloved-hand, and robot embodiments. On measured recordings, it improves temporal accuracy by 5.38 points and Volumetric IoU by 5.02 points under a shared protocol. These results show that heterogeneous human--robot supervision and privileged training views can turn egocentric video into dense tactile annotations.
\end{abstract}

\afterpage{%
\begin{figure}[!t]
\centering
\includegraphics[width=\linewidth]{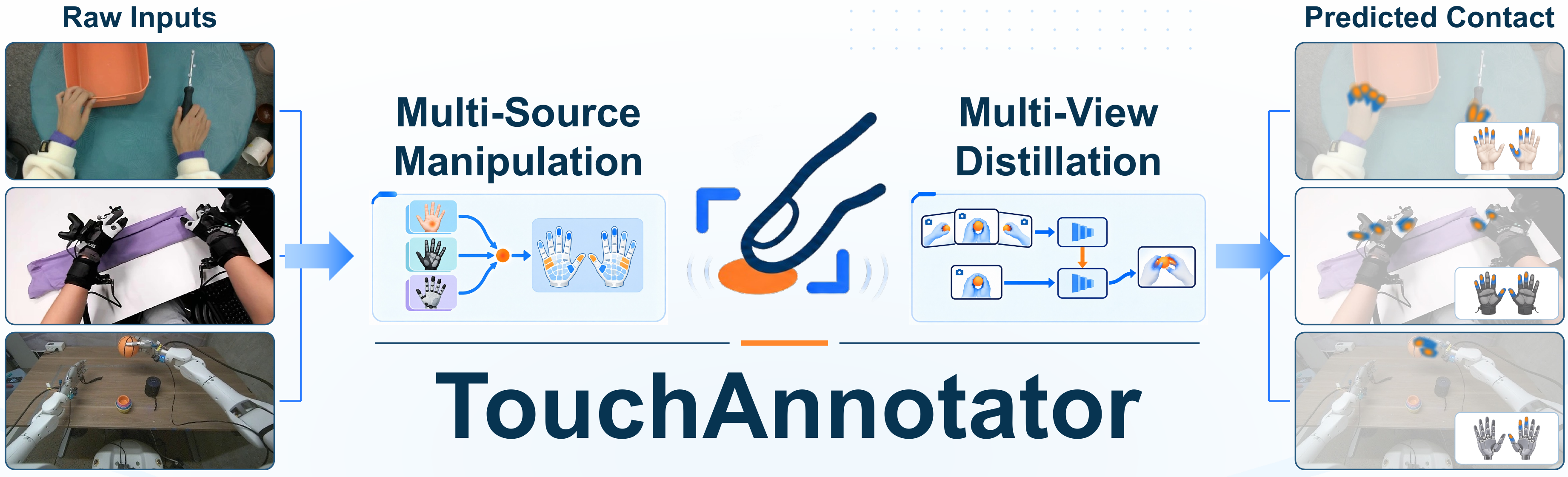}
\caption{\textbf{Tactile annotation from egocentric video.} Human and robot supervision supports bimanual relative-pressure and contact estimates from ego RGB and pose estimated from that stream. The heatmaps are schematic and do not depict predictions or sensor readings.}
\label{fig:Teaser}
\end{figure}%
}

\section{Introduction}
\label{sec:introduction}

Understanding manipulation requires knowing where the hands make contact and how that contact changes over time. Egocentric datasets capture diverse everyday activities and detailed hand--object motion~\citep{damen2018epic,grauman2022ego4d,liu2022hoi4d}, but these visual records do not directly measure touch. A video may show a hand closing around a bottle while leaving the contact regions hidden behind the object. Annotating existing videos with contact and relative-pressure maps could add a useful source of supervision for understanding human interaction and learning from demonstrations, extending the value of recordings collected without tactile sensors.

Vision-based tactile prediction offers a route toward this goal. PressureVision and EgoPressure infer pressure from visual observations of hand--surface interaction~\citep{grady2022pressurevision,zhao2025egopressure}; OpenTouch and EgoTactile extend tactile data collection to everyday object manipulation~\citep{song2025opentouch,zeng2026egotactile}; and EgoTac combines measured intensity with geometry-derived contact across human datasets~\citep{zhang2026egotac}. These sources provide complementary supervision but differ in sensor layout, spatial support, scale, embodiment, and whether they measure pressure, contact, or deformation. Moreover, an egocentric camera can miss the hand--object interface precisely when contact occurs, even when synchronized wrist views are available during data collection~\citep{zhou2026touchanything}. This raises a broader question: \emph{how can heterogeneous tactile sources and privileged training views be combined to train a single annotator that requires only egocentric input at deployment?}

Related advances in egocentric learning motivate treating auxiliary observations as privileged training information. Ego-Exo4D provides synchronized first- and third-person views of skilled activities~\citep{grauman2024egoexo4d}, while Ego-Exo transfers representations learned from third-person videos to egocentric recognition~\citep{li2021egoexo}. Multimodal distillation similarly transfers complementary training signals into an RGB-only action recognizer~\citep{radevski2023multimodal}. These approaches connect to the broader principle of learning with information available only during training~\citep{lopezpaz2016unifying}. Tactile annotation adds a second challenge: supervision from different sensors and embodiments must share an anatomical representation without treating missing regions or incompatible targets as negatives.

We introduce \method{}, a framework built around multi-source tactile supervision and privileged-view distillation (Figure~\ref{fig:Teaser}). First, a canonical bilateral hand field unifies glove pressure, geometry-derived human contact, and sparse robot fingertip deformation from six datasets and three embodiments. Source-aware masks preserve which hand regions and target types are valid, allowing heterogeneous signals to train a shared tactile representation without inventing supervision for unobserved cells.

Second, an exponential-moving-average teacher observes clean egocentric and synchronized wrist videos during training. An ego-only student learns from the teacher's hand-local representations and pressure/contact predictions while receiving degraded visual input. At deployment, the student uses only egocentric video and hand pose estimated from the same stream. A compact memory of preceding clips supplies temporal context, while the predicted intensity remains sensor-relative because visual observations do not uniquely determine physical pressure. We additionally collect egocentric video with synchronized glove measurements to validate predictions on everyday manipulation tasks.

Our contributions are threefold:
\begin{enumerate}
    \item \textbf{Multi-source tactile supervision}: a canonical bimanual field with source-aware masks unifies pressure, contact, and deformation signals from six datasets and three human--robot embodiments.
    \item \textbf{Privileged-view distillation}: an ego--wrist teacher transfers hand-local representations and tactile predictions to a student that requires only ego video and estimated pose at deployment.
    \item \textbf{Extensive validation in real-world settings}: experiments on EgoTouch and FPHA, together with an evaluation on our synchronized glove recordings, examine tactile prediction, cross-dataset transfer, and agreement with measured touch.
\end{enumerate}

Our experiments examine ego-only tactile prediction, cross-dataset transfer, and agreement with measured touch. On EgoTouch, the model predicts bimanual tactile fields from ego RGB and estimated hand pose, while additional visual evidence improves contact and intensity overlap for the same checkpoint. Without FPHA adaptation or threshold tuning, \method{} reaches F1 $0.3658$ on geometry-derived bare-hand contact. On 860 synchronized glove recordings, it improves temporal accuracy by $5.38$ points and Volumetric IoU by $5.02$ points under the shared sensor-grounded protocol. Qualitative examples further show task-relevant tactile responses across bare-hand, gloved-hand, and robot embodiments. These results support multi-source transfer and the use of privileged training views while keeping cross-protocol comparisons explicitly qualified.

\Needspace{6\baselineskip}
\section{Related work}
\label{sec:related}

\begin{table}[!t]
\caption{\textbf{Scale and diversity of the configured training corpus.} Counts cover unique segments before ratio sampling and native synchronized timesteps counted once across cameras. P., C., and Priv. denote relative pressure, contact, and non-ego teacher views; Geom. and Deform. denote geometry- and deformation-derived contact.}
\label{tab:data_sources}
\centering
\small
\setlength{\tabcolsep}{2pt}
\begin{tabular*}{\linewidth}{@{\extracolsep{\fill}}lrrrccc@{}}
\toprule
Source & Segments & Frames & Hours & P. & C. & Priv. \\
\midrule
\multicolumn{7}{@{}l}{\emph{Tactile gloves: measured sensor responses}} \\
EgoTouch~\citep{zhou2026touchanything} & 541 & 411,568 & 3.8 & \checkmark & \checkmark & \checkmark \\
HumanTouch~\citep{humantouch2026} & 17,545 & 28,505,945 & 132.0 & \checkmark & \checkmark & \checkmark \\
\midrule
\multicolumn{7}{@{}l}{\emph{Bare hands: geometry-derived contact}} \\
HOI4D~\citep{liu2022hoi4d} & 1,415 & 351,050 & 6.5 & -- & Geom. & -- \\
TACO~\citep{liu2024taco} & 2,311 & 361,631 & 3.3 & -- & Geom. & -- \\
OakInk2~\citep{zhan2024oakink2} & 822 & 1,004,594 & 9.3 & -- & Geom. & -- \\
\midrule
\multicolumn{7}{@{}l}{\emph{Dexterous robot hands: optical deformation}} \\
T-Rex~\citep{niu2026trex} & 4,371 & 4,363,007 & 40.4 & -- & Deform. & \checkmark \\
\midrule
\textbf{Total} & \textbf{27,005} & \textbf{34,997,795} & \textbf{195.3} & & & \\
\bottomrule
\end{tabular*}
\end{table}

\subsection{Egocentric interaction and tactile prediction}

Egocentric datasets span diverse activities~\citep{damen2018epic,grauman2022ego4d}, RGB-D hand motion~\citep{garciahernando2018fpha}, and detailed hand--object geometry~\citep{liu2022hoi4d,fan2023arctic,zhan2024oakink2,liu2024taco}. ContactDB records thermal grasp traces, while DECO and HACO estimate dense human--scene or hand contact from RGB~\citep{brahmbhatt2019contactdb,tripathi2023deco,jung2025haco}.

Body and hand reconstruction uses transformer-based mesh recovery~\citep{goel2023humans4d,pavlakos2024hamer,potamias2025wilor}, diffusion priors for occluded bodies~\citep{zhu2024dpmesh}, and task-aware 3D keypoint localization for close-range HRI~\citep{li2026taihri}. Joint human--object and hand--object reconstruction incorporates contact constraints~\citep{hasson2019joint,cao2021rho,xie2022chore,fan2024hold}, dense distance fields~\citep{fan2023arctic}, and generative priors~\citep{ye2024ghop,li2025scorehoi}. These methods estimate geometry and contact; pressure remains unmeasured.

Vision--touch learning connects cross-modal representations~\citep{li2019connecting,yang2022touchandgo} with visual pressure estimation~\citep{grady2022pressurevision,zhao2025egopressure}. OpenTouch, EgoTouch/TouchAnything, and EgoTactile extend tactile sensing and prediction to everyday manipulation~\citep{song2025opentouch,zhou2026touchanything,zeng2026egotactile}. TouchAnything supports ego-only inference through view dropout, while EgoTac combines measured intensity and geometry-derived contact with task-valid masks, dual heads, and short RGB histories~\citep{zhou2026touchanything,zhang2026egotac}. We emphasize privileged multi-view distillation, bounded memory across clips, and source-masked robot deformation supervision.

\subsection{Tactile policy learning and privileged supervision}

Vision--language--action (VLA) models connect pretrained visual--language knowledge to robot control. RT-2 represents actions as tokens~\citep{zitkovich2023rt2}, while $\pi_0$ uses flow matching for continuous actions~\citep{black2024pi0}. OpenVLA scales open-source policies across robot embodiments, while EgoForesight incorporates noisy egocentric actions and privileged-future distillation for dexterous policy learning~\citep{kim2025openvla,zhang2026egoforesight}. These models motivate tactile feedback for physical interaction.

Tactile-VLA combines VLA priors with tactile feedback and hybrid position--force control, while OmniVTLA aligns touch with vision and language~\citep{huang2025tactilevla,cheng2025omnivtla}. T-Rex separates visuomotor planning from high-rate tactile refinement~\citep{niu2026trex}. TTP transfers human tactile--action pretraining through shared human/robot spaces, and FTP-1 scales tactile pretraining across sensors and embodiments~\citep{zhang2026ttp,yuan2026ftp1}. Being-H0.8 combines pseudo-touch and measured touch in a latent world--action model; Motus2 refines action sub-chunks and predicts future force~\citep{beingbeyond2026beingh08,bi2026motus2}. Our dense, hand-aligned video annotations could supply tactile supervision for such policies; downstream control remains outside our evaluation.

Privileged learning and distillation transfer training-only information to a deployable student~\citep{vapnik2009lupi,hinton2015distilling,lopezpaz2016unifying}. Cross-modal feature transfer, modality hallucination, and multimodal distillation exploit auxiliary observations during training while reducing input requirements at inference~\citep{gupta2016crossmodal,hoffman2016hallucination,yuan2022xtrans2cap,radevski2023multimodal}. We apply this principle to hand-local features and dense pressure/contact fields, transferring synchronized wrist-view information to an ego-only student through feature matching and source-masked output distillation, with confidence- and discrepancy-based gating. Segment-recurrent memory~\citep{dai2019transformerxl} compresses past clips into tokens for current tactile queries.

\section{Method}
\label{sec:method}

\begin{figure}[!t]
    \centering
    \includegraphics[width=\linewidth]{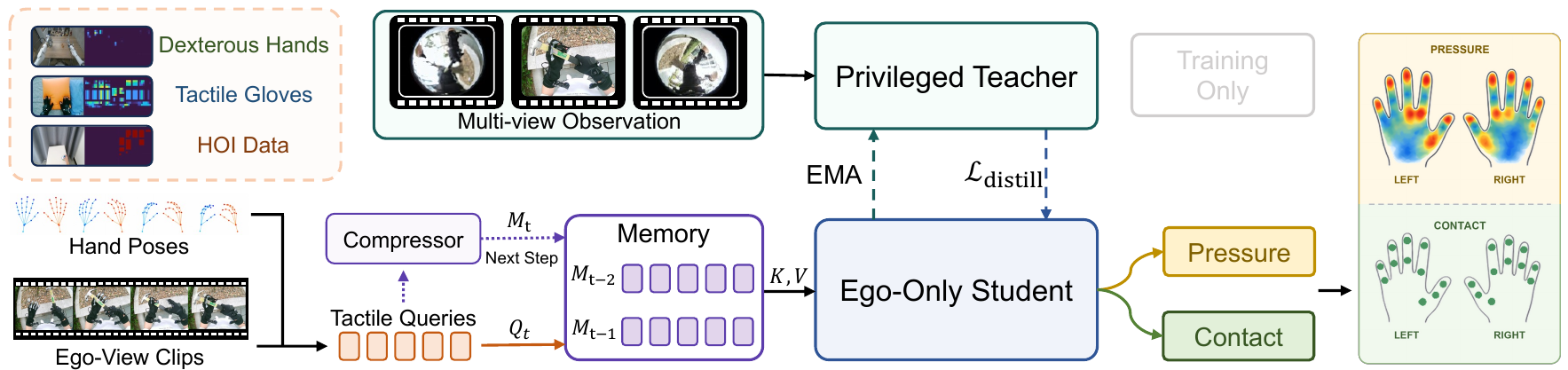}
    \caption{\textbf{Training and deployment of TouchAnnotator.} Human and robot supervision supplies tactile targets, and a multi-view EMA teacher guides the ego-only student during training. Compressed memory provides past-clip context for the student at deployment.}
    \label{fig:pipeline}
\end{figure}

\subsection{Problem formulation}

Index clips by $t$. Each contains $T$ egocentric frames $\mathbf{X}^{e}_{t}\in\mathbb{R}^{T\times 3\times H\times W}$ and bimanual pose $\mathbf{P}_{t}\in\mathbb{R}^{T\times 42\times 3}$. Training samples may also provide synchronized wrist clips $\mathbf{X}^{w_l}_{t},\mathbf{X}^{w_r}_{t}$. For each clip, the model predicts bimanual pressure and contact maps on a hand-aligned grid:
\begin{equation}
    \hat{\mathbf{Y}}^{p}_{t},\hat{\mathbf{Y}}^{c}_{t}
    \in [0,1]^{T\times 2\times G\times G}, \qquad G=21,
\end{equation}
where $p$ is relative pressure and $c$ is contact probability. A finite-cell mask $\mathbf{M}$ and sample-level availability flags $a^{p},a^{c}\in\{0,1\}$ define task masks $\mathbf{M}^{p}=a^{p}\mathbf{M}$ and $\mathbf{M}^{c}=a^{c}\mathbf{M}$. Deployment uses
\begin{equation}
  f_{\theta}:\big(\mathbf{X}^{e}_{t-K:t},\mathbf{P}_{t-K:t}\big)
  \mapsto \big(\hat{\mathbf{Y}}^{p}_{t},\hat{\mathbf{Y}}^{c}_{t}\big),
\end{equation}
with pose estimated from ego RGB using WiLoR in the reported evaluations. Ego-only deployment thus estimates pose without tactile measurements or wrist images.

\subsection{Unifying heterogeneous tactile supervision}
\label{sec:heterogeneous}

Our configured training corpus spans \textbf{six datasets and three embodiment types}, totaling \textbf{27,005 video segments and 35.0M synchronized source frames} (195.3 hours; Table~\ref{tab:data_sources}). It combines measured glove responses, geometry-derived human contact, and robot fingertip deformation. Only EgoTouch and HumanTouch supervise relative pressure; all six sources provide contact supervision.

All sources share a bilateral $21\times21$ anatomical layout. EgoTouch remaps each $16\times16$ glove array; HumanTouch projects 360 taxels onto the same 217-cell support and uses one dataset-level monotone map, fixed across clips, to match EgoTouch's active-taxel distribution. This statistical alignment does not establish common physical pressure units. Geometry-derived contact thresholds MANO vertex-object distances at 5 mm before interpolation to the grid; proximity remains a noisy contact proxy. T-Rex deformation is pooled into corresponding fingertip regions, leaving unobserved cells invalid. Hand mirroring follows a shared anatomical convention. The mixed loader uses fixed source proportions (Table~\ref{tab:run_audit}), avoiding repetition where possible.

\subsection{Hand-local tactile queries and dual prediction heads}

Each available stream yields local tokens $\mathbf{F}^{v}\in\mathbb{R}^{T\times N\times D}$ through a shared Qwen3.5 video tower. A pose encoder supplies hand-specific tokens. Each hand has 64 learned tactile queries on an $8\times8$ latent grid, attending to ego and its corresponding wrist stream; both use ego only at deployment.

A pose-conditioned soft ROI weights visual tokens at within-clip frame $\tau$:
\begin{equation}
 r^{h,v}_{\tau,n}=\sigma\!\left(
 \frac{\langle W_q\bar{\mathbf{p}}^{h}_{\tau},W_k\mathbf{F}^{v}_{\tau,n}\rangle}{\sqrt{D}}
 \right),
\end{equation}
where $\bar{\mathbf{p}}^{h}_{\tau}$ averages hand $h$'s pose tokens. Two cross-attention blocks fuse ROI-weighted features, pose, and tactile queries into $\mathbf{Q}_{t}\in\mathbb{R}^{T\times2\times64\times D}$. After memory augmentation (Section~\ref{sec:memory}), a Qwen3.5 reasoner produces $\mathbf{Z}_{t}$ using shared forward and reversed-time passes. Independent output projections predict pressure and contact, separating their logits while retaining a shared representation.

\subsection{Multi-view distillation}
\label{sec:distillation}

Multi-view distillation applies to samples with valid privileged views. A stop-gradient teacher, initialized from the student, receives clean ego and available wrist clips. We update its parameters by an exponential moving average:
\begin{equation}
    \bar\theta \leftarrow m\bar\theta + (1-m)\theta,
    \qquad m=0.996,
\end{equation}
 The student receives ego only, with stochastic downsampling and sensor noise sharing one degradation plan across current and historical clips. Single-view samples train without distillation.

For reasoned queries, a student predictor $g$ matches normalized teacher features:
\begin{equation}
 \mathcal{L}_{q}=\frac{1}{|\Omega_q|}
 \sum_{i\in\Omega_q}\left(2-2\left\langle
 \frac{g(\mathbf{Z}^{s}_{i})}{\|g(\mathbf{Z}^{s}_{i})\|_2},
 \frac{\mathbf{Z}^{t}_{i}}{\|\mathbf{Z}^{t}_{i}\|_2}
 \right\rangle\right).
\end{equation}
Here $\Omega_q$ contains queries from privileged samples whose mean $1-\cos(\mathbf{Z}^{s},\mathbf{Z}^{t})$, computed before $g$, lies in $[0.002,0.50]$. Pressure transfer uses smooth-$\ell_1$ regression under this sample gate and $\mathbf{M}^{p}$. Contact transfer uses Bernoulli Jensen--Shannon divergence under $\mathbf{M}^{c}$, an independent sample-mean JSD gate $[10^{-5},0.10]$, and cellwise teacher confidence of at least $0.60$ in either class. These gates heuristically select moderate discrepancies; they do not guarantee teacher correctness. Empty masks yield zero loss. The combined term is
\begin{equation}
 \mathcal{L}_{\mathrm{distill}}=
 \mathcal{L}_{q}+0.5\mathcal{L}_{p}^{\mathrm{distill}}+
 0.5\mathcal{L}_{c}^{\mathrm{distill}}.
\end{equation}
Output distillation respects task availability; query distillation has no cell-level mask.

\subsection{Past-only causal clip memory}
\label{sec:memory}

The current queries read the $K=2$ preceding non-overlapping clips. For each, $M=16$ learned seeds cross-attend to queries with sinusoidal time encoding across frames, hands, and sites, producing $\mathbf{C}_{t-k}\in\mathbb{R}^{M\times D}$. Two cross-attention/MLP layers read recency-encoded caches:

\begin{equation}
 \widetilde{\mathbf{Q}}_{t}=
 \operatorname{MemRead}\!\left(
 \mathbf{Q}_{t},
 [\mathbf{C}_{t-2}+\mathbf{e}_{2};\mathbf{C}_{t-1}+\mathbf{e}_{1}]
 \right).
\end{equation}
Missing history is masked. Each cache compresses its clip's unconditioned queries, excluding memory readouts and temporal-reasoner outputs, so history does not accumulate recursively. The state is bounded to two caches. Causality holds \emph{across clips}; bidirectional reasoning requires observing the complete current eight-frame clip and is not frame-online inference.

\subsection{Training objective}

In the saved Full configuration, $\mathcal{L}_{\mathrm{press}}$ combines contact-weighted MSE and $\ell_1$ regression with total variation under $\mathbf{M}^{p}$; $\mathcal{L}_{\mathrm{contact}}$ is positive-weighted BCE under $\mathbf{M}^{c}$. The objective is
\begin{equation}
 \mathcal{L}=
 \mathcal{L}_{\mathrm{press}}+
 \lambda_c\mathcal{L}_{\mathrm{contact}}+
 \lambda_{\mathrm{roi}}\mathcal{L}_{\mathrm{roi}}+
 \lambda_{\mathrm{align}}\mathcal{L}_{\mathrm{align}}+
 \lambda_{\mathrm{distill}}\mathcal{L}_{\mathrm{distill}}.
\end{equation}
ROI supervision uses available glove masks. Alignment couples visual, pose, fused, and tactile embeddings; its separate tactile encoder reconstructs source signals on valid cells, including contact proxies. The distillation weight ramps as $\lambda_{\mathrm{distill}}=0.1\min(1,e/3)$ at epoch $e\geq1$. Settings appear in Appendix~\ref{app:implementation}.

\section{Experiments}
\label{sec:experiments}

\afterpage{%
\begin{figure}[!t]
\centering
\includegraphics[width=\linewidth]{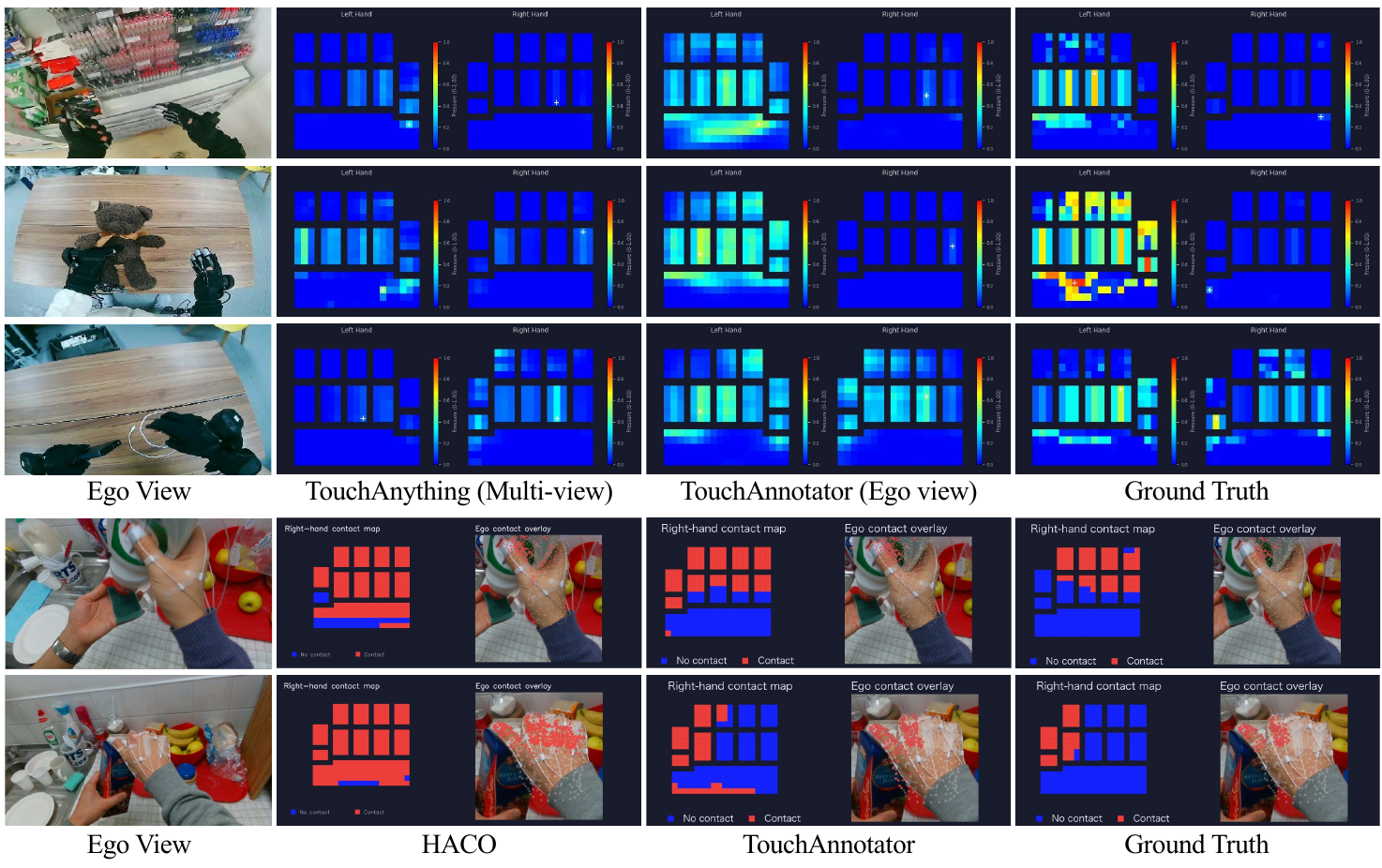}
\caption{\textbf{Qualitative tactile predictions on EgoTouch and FPHA.} The upper examples compare EgoTouch pressure estimates from multi-view TouchAnything and ego-only TouchAnnotator with measured maps. The lower examples compare FPHA contact estimates from HACO and TouchAnnotator with geometry-derived reference labels.}
\label{fig:main_vis}
\end{figure}%
}

\begin{table*}[!t]
\caption{\textbf{Tactile prediction on EgoTouch.} Local TouchAnnotator results use 184 seen and 76 unseen manifest-A episodes; all-view input adds both wrist cameras. TouchAnything rows are published references whose split equivalence to the local evaluations is unverified.}
\label{tab:egotouch_main}

\centering
\small
\resizebox{\textwidth}{!}{%
\begin{tabular}{lcccccccc}
\toprule
& \multicolumn{4}{c}{Seen objects}
& \multicolumn{4}{c}{Unseen objects} \\
\cmidrule(lr){2-5}
\cmidrule(lr){6-9}
Method
& T.Acc.$\uparrow$
& C.IoU$\uparrow$
& V.IoU$\uparrow$
& MAE$\downarrow$
& T.Acc.$\uparrow$
& C.IoU$\uparrow$
& V.IoU$\uparrow$
& MAE$\downarrow$ \\
\midrule

\multicolumn{9}{l}{\emph{Multi-view}} \\
TouchAnything~\citep{zhou2026touchanything}
& 0.8566 & 0.5030 & 0.4575 & 0.0436
& 0.8347 & 0.4496 & 0.3852 & 0.0601 \\
\method{} student (all views)
& \textbf{0.8713} & \textbf{0.5139} & \textbf{0.4578} & 0.0471
& \textbf{0.9127} & \textbf{0.5038} & \textbf{0.4061} & 0.0652 \\

\midrule
\multicolumn{9}{l}{\emph{Egocentric}} \\
TouchAnything~\citep{zhou2026touchanything}
& 0.8393 & 0.4792 & 0.4311 & 0.0456
& 0.8271 & 0.4396 & 0.3743 & 0.0615 \\
\method{} student (ego only)
& \textbf{0.8613} & \textbf{0.5051} & \textbf{0.4478} & 0.0479
& \textbf{0.8687} & \textbf{0.4779} & \textbf{0.3821} & 0.0638 \\
\bottomrule
\end{tabular}%
}
\end{table*}

We first evaluate on the established EgoTouch and FPHA benchmarks and compare qualitative predictions on both datasets. We then test agreement with measured touch on our collected recordings and examine whether a shared canonical field supports qualitatively coherent predictions across embodiments.

\subsection{Experimental setup}

EgoTouch results use ego-estimated WiLoR pose~\citep{potamias2025wilor} and episode-macro metrics on 184 seen and 76 unseen manifest-A episodes. Contact IoU thresholds pressure at $0.1$; the published TouchAnything scores are a reference because split equivalence is unverified. FPHA groups RGB-only baselines separately from methods given estimated hand pose. The sensor-grounded comparison uses TouchAnything on measured cells. We report overlap and absolute-error metrics together to distinguish spatial contact coverage from pressure calibration. Appendix~\ref{app:experimental_setup} documents the evaluation protocols and checkpoints.

\subsection{Main results on EgoTouch}

Table~\ref{tab:egotouch_main} juxtaposes the local Full seen/unseen evaluations with a published reference whose split equivalence is unverified; it supports no SOTA margin. On the same 76 episodes, providing all views increases Contact IoU from $0.4779$ to $0.5038$ and Volumetric IoU from $0.3821$ to $0.4061$, while MAE rises from $0.0638$ to $0.0652$. A similar trade-off appears against TouchAnything: \method{} achieves higher overlap but slightly higher MAE. This is consistent with sparse tactile maps: MAE over all cells favors conservative near-zero predictions, whereas our positive-pressure weighting and contact supervision encourage broader active regions. The latter improves contact coverage but can accumulate small background residuals. On seen objects, additional views instead reduce MAE from $0.0479$ to $0.0471$, suggesting better pressure calibration for familiar objects. Both input conditions use the same checkpoint, so this comparison isolates inference-time view access.

The all-view result is a diagnostic rather than the deployment setting. Because both rows use the same student weights and ego-estimated pose, the difference measures the value of additional image evidence without changing the learned predictor or pose source. The remaining gap indicates that wrist cameras reveal contact cues hidden from the head-mounted view, while the ego-only row represents the intended annotation protocol for videos that lack synchronized auxiliary cameras.

\subsection{Transfer to bare-hand contact on FPHA}

FPHA tests whether the tactile predictor transfers to contact annotations on bare hands. We evaluate the 124 sequences (10,998 frames) in its official test split with complete hand/object geometry~\citep{garciahernando2018fpha}. Every method uses the same cached labels: a MANO hand vertex is in contact when its nearest object-mesh vertex is within 10 mm. Predictions at shared 15 Hz timestamps are linearly interpolated to the 30 Hz source grid; metrics are computed over 778 right-hand vertices per timestamp and then averaged. We perform no FPHA training, threshold tuning, or label-dependent mapping fit.

Table~\ref{tab:fpha} separates the RGB-only methods DECO~\citep{tripathi2023deco} and HACO~\citep{jung2025haco} from TouchAnything and \method{}, which also receive estimated 3D right-hand joints. DECO's full-body scores are restricted to the same right-hand vertex subset, HACO predicts native MANO scores, and both pressure-based models use the same fixed 217-to-778 nearest-support projection. Binary contact uses DECO probability $\geq0.5$, HACO probability $>0.52$, or mapped pressure $>0.1$; AUROC uses continuous scores.

Without FPHA adaptation, \method{} reaches F1 $0.3658$ on these bare-hand labels, close to HACO's $0.3768$. Its AUROC ($0.7431$ versus $0.7376$) and recall ($0.6169$ versus $0.5742$) are also numerically higher. Under the shared estimated-pose protocol, its F1 exceeds TouchAnything's $0.1209$. This shows that a predictor trained with heterogeneous tactile supervision can remain competitive on geometry-derived bare-hand contact. Figure~\ref{fig:main_vis} shows representative EgoTouch pressure and FPHA contact maps. Appendix~\ref{app:fpha_protocol} details preprocessing, timestamp handling, and mapping sensitivity.

The input groups in Table~\ref{tab:fpha} should be interpreted separately: the pose-conditioned models receive information unavailable to the RGB-only baselines. Within this protocol, AUROC evaluates score ranking independently of a single operating threshold, whereas F1 reflects the fixed threshold used by each output type. \method{}'s higher AUROC and recall but slightly lower F1 than HACO therefore indicate transferable contact ranking without establishing a universally better calibrated decision rule.

\begin{table}[!t]
\caption{\textbf{Cross-dataset contact transfer on FPHA.} Methods share geometry-derived labels and timestamps on 124 official-test sequences. Groups distinguish ego RGB alone from ego RGB with estimated hand pose; bold marks the best result within each group.}
\label{tab:fpha}
\centering
\small
\setlength{\tabcolsep}{8pt}
\begin{tabular}{lccc}
\toprule
Method & AUROC$\uparrow$ & Recall$\uparrow$ & F1$\uparrow$ \\
\midrule
\multicolumn{4}{l}{\emph{Ego RGB only}} \\
DECO & 0.6763 & 0.1550 & 0.0858 \\
HACO & 0.7376 & 0.5742 & \textbf{0.3768} \\
\midrule
\multicolumn{4}{l}{\emph{Ego RGB + estimated hand pose}} \\
TouchAnything & 0.5486 & 0.1153 & 0.1209 \\
\method{} & \textbf{0.7431} & \textbf{0.6169} & 0.3658 \\
\bottomrule
\end{tabular}
\end{table}

\subsection{Validation on collected tactile recordings}
\label{sec:benchmark}

\begin{figure}[!t]
\centering
\includegraphics[width=\linewidth]{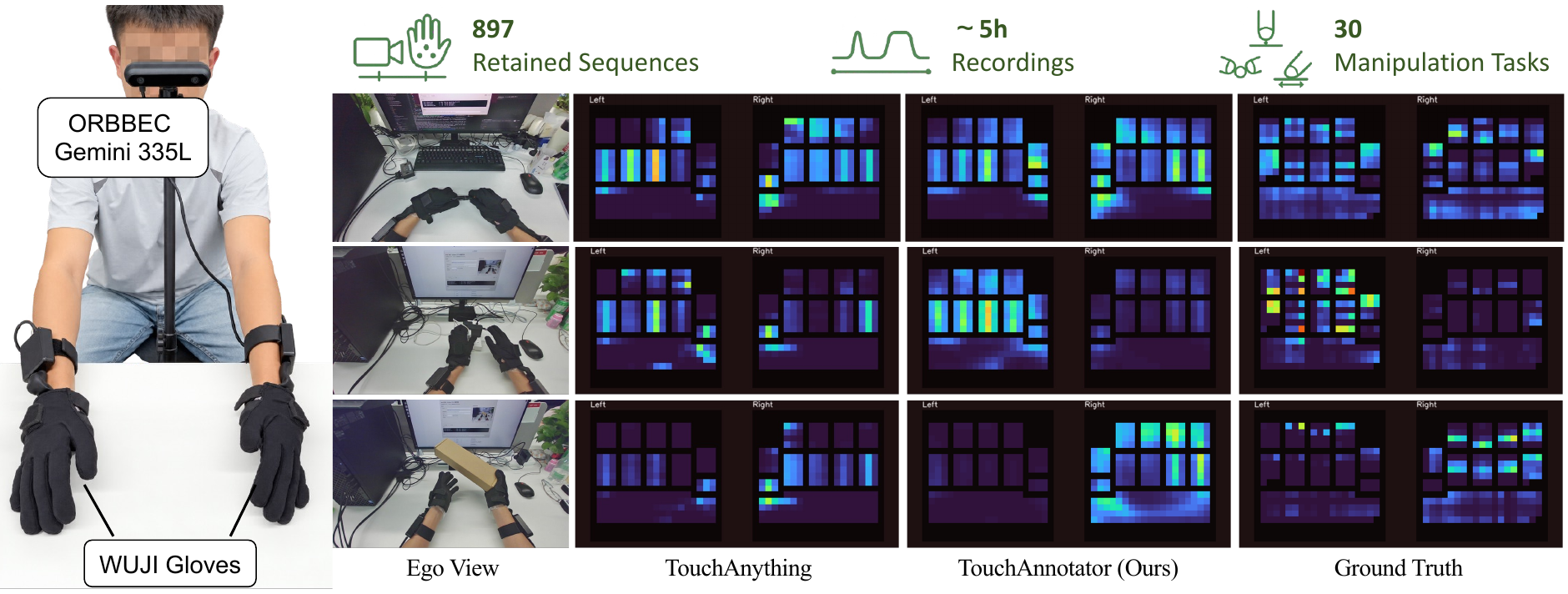}
\caption{\textbf{Collected recordings and measured tactile examples.} An ego camera and Wuji gloves record synchronized bimanual interactions; the panels compare example predictions with measured maps. The collection contains 897 retained sequences across 30 tasks (about five hours).}
\label{fig:data_collection}
\end{figure}

For additional validation, we collected egocentric video with synchronized tactile measurements from both hands: \textbf{897 retained sequences across 30 everyday manipulation tasks} (Figure~\ref{fig:data_collection}). These recordings provide a measured reference for contact location and relative-intensity changes when visible motion alone may be ambiguous. Appendix Table~\ref{tab:benchmark_tasks} defines every task and its principal tactile events.

\paragraph{Diverse contact mechanisms.}
The tasks span brief presses, insertion and separation, sustained grasping, and wiping. This coverage supports evaluation of contact onset and release, changes in active hand regions, and intensity variation during continued interaction. It includes simple actions with brief tactile events as well as manipulation involving sustained or changing contact constraints.

\paragraph{Bimanual coordination and temporal structure.}
Joint support, handovers, and complementary hand roles extend evaluation to contact distribution across both hands. The synchronized streams allow predictions to be checked for coordinated contact changes and continuity during stable manipulation. These interactions motivate temporal diagnostics for brief events, delayed transitions, and spurious changes during sustained contact.

\paragraph{Evaluation protocol.}
The annotator receives ego RGB and hand pose estimated from that stream. Evaluation uses common measured hand regions and trajectory-relative sensor responses. Contact IoU assesses spatial contact, Volumetric IoU and MAE assess relative intensity, and temporal accuracy and F1 assess frame-level contact presence. Alignment, normalization, and metric definitions appear in Appendix~\ref{app:benchmark_protocol}. Table~\ref{tab:bench_main} reports a cached raw 860-episode cohort; it does not cover all 897 retained sequences in the collection.

\begin{table}[!t]
\caption{\textbf{Sensor-grounded tactile prediction by expected tactile value.} Each cell compares TouchAnything / TouchAnnotator on a cached cohort of 860 episodes from 21 tasks using measured-support 217-cell split-pressure targets. Lower to High are preliminary labels based on task descriptions; the operation-complexity breakdown appears in Table~\ref{tab:bench_complexity}.}
\label{tab:bench_main}
\centering
\footnotesize
\setlength{\tabcolsep}{3pt}
\renewcommand{\arraystretch}{1.06}
\begin{tabular*}{\linewidth}{@{\extracolsep{\fill}}lcccc@{}}
\toprule
Metric (TA / Ours) & Overall (21/860) & Lower (3/110) & Moderate (11/430) & High (7/320) \\
\midrule
T.Acc.$\uparrow$ & 63.29/\textbf{68.67} & 65.12/\textbf{72.41} & 71.59/\textbf{75.31} & 48.20/\textbf{55.66} \\
T.F1$\uparrow$ & 76.87/\textbf{81.08} & 78.86/\textbf{83.99} & 82.61/\textbf{85.47} & 65.02/\textbf{71.46} \\
C.IoU$\uparrow$ & 6.89/\textbf{10.73} & 7.11/\textbf{10.11} & 8.04/\textbf{11.76} & 5.21/\textbf{9.38} \\
V.IoU$\uparrow$ & 9.10/\textbf{14.12} & 9.23/\textbf{13.97} & 10.37/\textbf{14.99} & 7.30/\textbf{12.83} \\
MAE$\downarrow$ & \textbf{0.083013}/0.087488 & \textbf{0.095079}/0.106340 & \textbf{0.073978}/0.081586 & 0.093117/\textbf{0.089496} \\
\bottomrule
\end{tabular*}
\end{table}

Table~\ref{tab:bench_main} shows that \method{} agrees more closely with measured tactile signals than TouchAnything. Overall, it improves temporal accuracy and F1 by 5.38 and 4.21 percentage points, respectively, while raising Contact IoU from 6.89 to 10.73 and Volumetric IoU from 9.10 to 14.12. The gains are consistent across tactile-value strata: \method{} improves both overlap metrics in every group and delivers the largest temporal gains in the high-value group, where temporal accuracy and F1 increase by 7.46 and 6.44 points. Although its overall MAE is slightly higher, it reduces MAE in the high-value group (0.0931 to 0.0895), indicating a clear advantage on interactions for which tactile information is most important. Additional breakdowns and evaluation details are provided in the appendix.

This combination of overlap gains and overall MAE regression reflects the different emphasis of the metrics. Contact and Volumetric IoU focus on recovering active support and its magnitude, whereas MAE averages errors over all valid cells, including the many inactive or weakly active locations. Reporting them together prevents a sparse near-zero prediction from appearing uniformly preferable. The tactile-value strata are descriptive task groupings rather than sensor-derived classes, so they reveal where the aggregate differences occur without defining a new supervised label.

\subsection{Cross-embodiment examples}

\begin{figure}[!t]
\centering
\includegraphics[width=\linewidth]{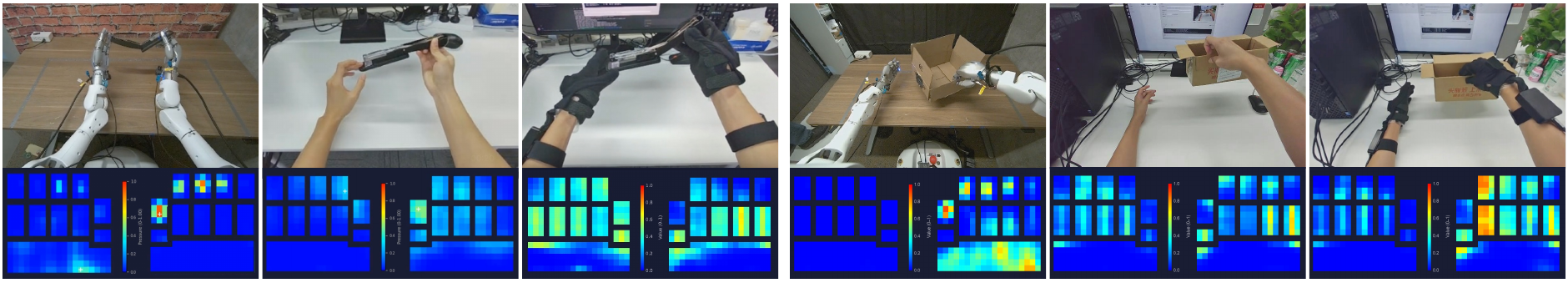}
\caption{\textbf{Illustrative tactile patterns across embodiments.} Two examples pair camera views with estimated tactile maps from dexterous robot hands, bare hands, and tactile gloves. }
\label{fig:cross_embodiment}
\end{figure}

Figure~\ref{fig:cross_embodiment} compares matched manipulation tasks performed by dexterous robot hands, bare hands, and gloved hands. For the same operation, \method{} predicts localized tactile responses across all three embodiments despite substantial differences in hand appearance and sensor layout. These examples qualitatively show that the learned representation can recover task-relevant touch across embodiments. Because the robot and glove responses have different sensor semantics and scales, the figure demonstrates cross-embodiment prediction rather than numerical alignment.

The canonical field makes these outputs visually comparable by anatomical region, not by physical unit. Accordingly, the figure is evidence of qualitative localization across appearance and embodiment shifts; it does not claim that equal colors correspond to equal forces. Establishing such a correspondence would require paired calibration across glove taxels, bare-hand contact proxies, and robot deformation sensors.

\section{Conclusion}

\method{} combines masked human/robot supervision, multi-view distillation, and past-only clip memory for bimanual tactile annotation from egocentric video. EgoTouch results show improved contact overlap when the same student receives wrist views. The cached sensor-grounded cohort shows higher contact overlap than TouchAnything, and FPHA shows stronger contact transfer under shared annotated-pose input. Together, these evaluations support dense tactile annotation from ordinary ego video while distinguishing deployment results from diagnostic settings with additional inputs.

The predictions remain relative sensor responses rather than calibrated forces. Geometry-derived contact depends on hand/object reconstruction and distance thresholds, and ego-only inference remains sensitive to pose failure, blur, truncation, and prolonged occlusion. Causality is enforced across clips, but the complete current clip is processed bidirectionally, so the system is intended for offline annotation rather than frame-online feedback. Evaluating whether the resulting tactile annotations improve downstream policy or representation learning is an important next step.

\section*{Reproducibility statement}
Appendix~\ref{app:implementation} records configuration, conversion, and evaluation details. Current audits cannot verify the published EgoTouch split or every historical initialization.

\section*{Ethics statement}
The collected recordings contain egocentric human interaction and therefore may include privacy-sensitive visual information. Data collection should use informed consent, minimize the capture of bystanders and personally identifying content, and provide a deletion mechanism. Tactile values are normalized sensor responses unless a sensor is explicitly force-calibrated; they should not be interpreted as safety-certified force estimates.

\bibliography{references}
\bibliographystyle{iclr2027_conference}

\clearpage
\appendix
\numberwithin{figure}{section}
\numberwithin{table}{section}
\numberwithin{equation}{section}
\section{Implementation details}
\label{app:implementation}

\subsection{Repository-derived configuration}

Table~\ref{tab:implementation} summarizes the repository's resolved full configuration with causal memory and multi-view distillation. Evaluated checkpoint epochs and training mixtures are reported separately in Table~\ref{tab:run_audit}. Memory is causal across clips; processing within the current eight-frame clip is bidirectional.

\subsection{Gating for multi-view distillation}

The query gate accepts discrepancies in $[0.002,0.50]$; the contact gate accepts Bernoulli JSD in $[10^{-5},0.10]$ and requires teacher confidence of at least $0.60$. These are implementation thresholds; their sensitivity has not been evaluated.

\subsection{Pressure and contact losses}

The saved Full configuration weights pressure MSE, $\ell_1$, and total variation by $1.0$, $0.5$, and $0.01$, with positive-pressure-cell weight $3.0$. Contact uses BCE with positive-class weight $3.0$ and threshold $0.1$. Optional metric-MAE and pressure/contact soft-IoU losses are unspecified and default to zero. Overall pressure and contact weights are both $1.0$; ROI and alignment weights are $0.25$ and $0.10$. These are the audited continuation settings, distinct from earlier pressure-focused pretraining recipes. Language/vision learning rates are $10^{-5}$ and the adapter learning rate is $5\times10^{-5}$.

\section{Data conversion details}

\paragraph{Geometry-derived contact.}
For HOI4D, MANO vertices and object meshes are reconstructed in a shared camera frame; a vertex is in contact when its closest surface distance is at most 5 mm. TACO and OakInk2 use the same contract with their official poses and transforms. Contact is interpolated to valid canonical-grid cells, while pressure remains explicitly unavailable.

\paragraph{HumanTouch.}
The adapter aligns RGB with sidecar tactile records. Palm and finger regions are projected separately into the EgoTouch mask, then a dataset-level dead zone and monotone active-taxel quantile map align distributions without per-clip normalization. Invalid sensors remain NaN. A calibration-fit manifest disjoint from evaluation data has not been verified.

\paragraph{T-Rex.}
The robot adapter aligns head RGB, MANO-semantic kinematics, and ten fingertip deformation streams. Each fingertip is pooled to a $4\times3$ grid patch; unobserved cells are invalid. Deformation supervises contact but is not treated as calibrated pressure.

\FloatBarrier

\begin{table}[!ht]
\renewcommand{\thetable}{A.\arabic{table}}
\caption{\textbf{Implementation configuration.} Settings describe the full model with multi-view distillation and two-clip causal memory.}
\label{tab:implementation}
\centering
\small
\begin{tabular}{ll}
\toprule
Component & Setting \\
\midrule
Input & 8 frames, interval 2, target 15 Hz, $224\times224$ \\
Visual backbone & Qwen3.5-4B native video tower \\
Query width & 768 \\
Tactile queries & 64/hand, $8\times8$ latent grid \\
Query fusion & 2 layers, 8 heads \\
Output & 2 hands $\times$ $21\times21$, pressure/contact heads \\
Historical clips & 2 non-overlapping clips \\
Memory & 16 tokens/clip, 2 layers, 8 heads \\
EMA momentum & 0.996 \\
Student view & Ego only for privileged samples \\
Student degradation & Shared resize/noise plan for current/history \\
Distillation weights & Query 1.0, pressure 0.5, contact 0.5 \\
Distillation total weight & Linear ramp to 0.10 over epochs 1--3 \\
Training budget & Full: 30 epochs; controls: 15; AdamW, bfloat16 \\
\bottomrule
\end{tabular}
\end{table}

\section{Evaluation audit and additional experiments}
\label{app:ablation_protocol}

\subsection{Experimental setup}
\label{app:experimental_setup}

\paragraph{Datasets.}
EgoTouch contains 1,891 episodes over 208 tasks, approximately 2.1M synchronized frames, three RGB views, 42-joint bimanual pose, and bilateral pressure~\citep{zhou2026touchanything}. The five ablation configurations are evaluated on the same unseen subset of 40 episodes from four tasks (37,308 frames). Historical training and validation data, including initialization data, are not fully audited for all checkpoints. The seen evaluation contains 184 episodes from 102 tasks (219,104 frames), after excluding one of the original 185 episodes whose task is absent from the archived training split. Its episodes do not overlap the archived train, validation, or unseen sets; all its tasks occur in training. Table~\ref{tab:run_audit} records the actual ablation mixtures.

\paragraph{Baselines.}
The EgoTouch table includes published TouchAnything ego-only and three-view results~\citep{zhou2026touchanything}; split equivalence to our local evaluations has not been verified. The cached sensor-grounded comparison uses TouchAnything under the measured-cell metric protocol. On FPHA, we evaluate DECO and HACO from RGB alone, and TouchAnything and \method{} with annotated hand pose.

\paragraph{Cross-dataset protocol scope.}
Published EgoTac FPHA results~\citep{zhang2026egotac} use a different protocol and are therefore excluded from the common-protocol comparison. Training sources, including OakInk2 for the Full model, are not treated as unseen-dataset tests.

\paragraph{Implementation.}
Inputs contain eight frames at 15 Hz, resized to $224\times224$; Table~\ref{tab:implementation} gives architecture and optimization settings. The archived EgoTouch evaluator advances clip starts by one source frame and averages overlapping predictions, so the ablation scores include offline aggregation. All ablation configurations allow 15 epochs; their evaluated best checkpoints occur at epochs 1--3. In runs with multi-view distillation, the distillation weight ramps linearly to $0.1$ over epochs 1--3 and is already nonzero at the evaluated epochs.

\paragraph{Metrics.}
All five ablation reports use WiLoR pose and episode-macro aggregation. Contact IoU is computed from pressure maps thresholded at $0.1$, rather than the independent contact head. Temporal accuracy measures frame-level contact-state agreement and can be inflated by class imbalance; it does not measure transition timing. MAE and Volumetric IoU concern normalized pressure. The archived evaluator preserves pre-normalized targets, otherwise dividing by the trajectory's maximum ground-truth pressure across both hands (with denominator at least 1). The reports do not record which normalization branch each file used, limiting cross-protocol pressure comparisons. FPHA uses timestamp-macro vertex-contact metrics, with input conditions in Table~\ref{tab:fpha} and further protocol details in Appendix~\ref{app:fpha_protocol}.

\subsection{Available reports and training configurations}

The five ablation reports contain four-decimal per-episode metrics and episode-macro summaries for the same 40 episodes. Every evaluated frame has a finite prediction; this does not establish pose validity. The evaluator loads student weights (\texttt{model\_state\_dict}) and derives Contact IoU from thresholded pressure, so these exports assess neither the EMA teacher nor the independent contact head.

The two additional seen reports evaluate the same Full student with ego/all input on 184 episodes (102 tasks; 219,104 frames). The saved manifest excludes the single \texttt{pick\_up\_biscuits} episode because its task is absent from the archived training split. Both reports exactly match the selected trajectory list; their episode-macro summaries agree with means recomputed from the rounded episode rows. This split audit does not verify the historical initialization manifests or equivalence to the published split. In Table~\ref{tab:run_audit}, ET, HT, HOI, Oak, and TR denote EgoTouch, HumanTouch, HOI4D, OakInk2, and T-Rex; $\lambda_{\mathrm{distill}}$ is the target weight ramped linearly over epochs 1--3.

\begin{table}[!t]
\caption{\textbf{Training mixtures and evaluated checkpoints for the ablation runs.} Sampling ratios come from saved run configurations, and epochs from evaluated checkpoint metadata.}
\label{tab:run_audit}
\centering
\small
\resizebox{\linewidth}{!}{%
\begin{tabular}{lccccccccc}
\toprule
Run & ET & HT & HOI & TACO & Oak & TR & $\lambda_{\mathrm{distill}}$ & Best epoch & Epoch budget \\
\midrule
ET+HOI & .50 & 0 & .50 & 0 & 0 & 0 & .10 & 1 & 15 \\
No T-Rex & .444444 & .111111 & .166667 & .166667 & .111111 & 0 & .10 & 2 & 15 \\
w/o distillation + data change & .40 & 0 & .15 & .15 & .15 & .15 & 0 & 2 & 15 \\
Clean ego-only EMA teacher & .40 & .10 & .15 & .15 & .10 & .10 & .10 & 1 & 15 \\
Memory on (pre-memory init.) & .40 & .10 & .15 & .15 & .10 & .10 & .10 & 3 & 15 \\
\bottomrule
\end{tabular}}
\end{table}

The ego-only EMA and memory-on rows use best checkpoints from epochs 1 and 3, respectively. The ego-only EMA run initializes from $C_{\mathrm{mem}}$, whereas the memory-on run initializes from an earlier pressure-pretraining checkpoint, so they do not form a controlled teacher-view or memory pair. Their reports use the same 40 episode identifiers and per-episode frame counts as the other ablations; the two evaluation data paths resolve to the same directory and split-file SHA-256.

\begin{table}[!t]
\caption{\textbf{Per-hand pressure error for the ablation runs.} MAE is averaged equally over the same 40 unseen episodes.}
\label{tab:hand_mae}
\centering
\small
\begin{tabular}{lccc}
\toprule
Checkpoint & MAE$\downarrow$ & Left MAE$\downarrow$ & Right MAE$\downarrow$ \\
\midrule
ET+HOI & .0488 & .0515 & .0462 \\
No T-Rex & .0471 & .0503 & .0438 \\
w/o distillation + data change & .0457 & .0498 & .0417 \\
Clean ego-only EMA teacher & .0475 & .0510 & .0439 \\
Memory on (pre-memory init.) & .0600 & .0638 & .0562 \\
\bottomrule
\end{tabular}
\end{table}

\subsection{Full-model input-view diagnostic}

\begin{figure}[!t]
\centering
\includegraphics[width=\linewidth]{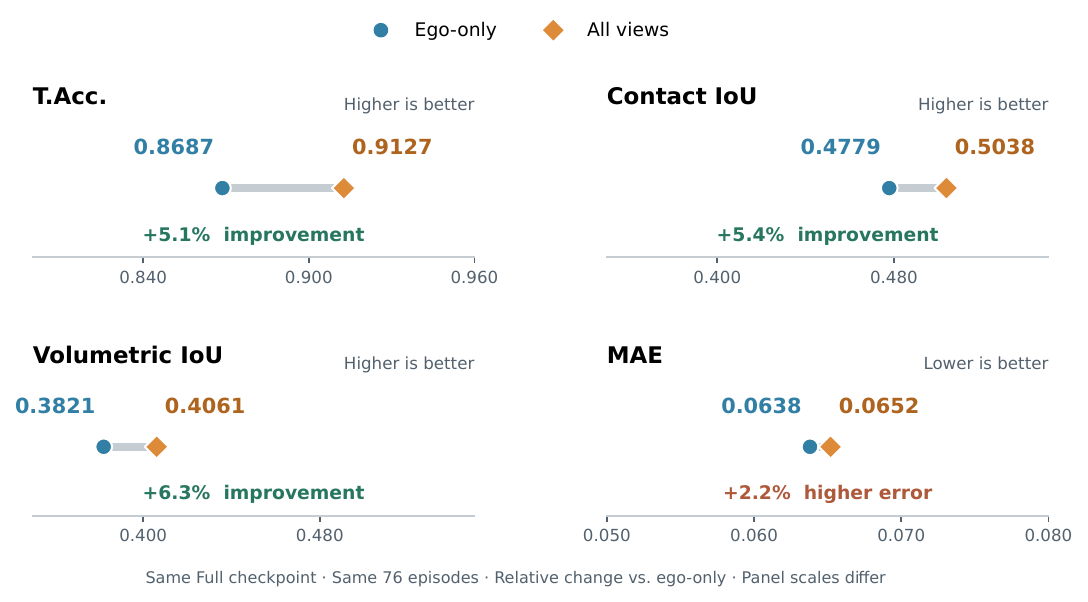}
\caption{\textbf{Effect of inference views.} Blue circles denote ego-only and orange diamonds all-view episode means for one Full checkpoint on 76 unseen episodes.}
\label{fig:full_view_differences}
\end{figure}

The paired ego/all reports use the same Full epoch-2 student, WiLoR pose, and 76 episodes with identical frame counts (Figure~\ref{fig:full_view_differences}). Mean Contact IoU increases by $0.0260$ (95\% episode-bootstrap interval $[0.0199,0.0325]$), improving on 68/76 episodes. Mean MAE increases by $0.0014$ despite decreasing on 47/76 episodes, indicating a smaller number of larger errors. Intervals use 20,000 paired episode resamples (seed 20260916) of the four-decimal exports; they condition on the observed task mixture and exclude unseen-task and training-seed uncertainty. Figure labels report $100(\mathrm{All}/\mathrm{Ego}-1)$; panel scales differ, with a shared scale only for the two IoU panels. Because both conditions use the same trained student, this experiment diagnoses inference-time view access rather than the effect of multi-view distillation.

\subsection{Task-stratified ablation results}

Table~\ref{tab:ablation_tasks} reports each task separately, preserving the episode weighting within each task. Rows are computed from four-decimal per-episode exports, so rounding can differ by $10^{-4}$ from full-precision recomputation; ``Move toolbox'' contributes 20 of the 40 episodes. Checkpoint and data differences in Table~\ref{tab:run_audit} still apply. The source reports provide task and trajectory identifiers but no independent participant/session identifiers for hierarchical uncertainty estimates.

\begin{table}[!t]
\caption{\textbf{Task-stratified ablation results.} Rows average exported per-episode scores within each of the four unseen evaluation tasks.}
\label{tab:ablation_tasks}
\centering
\small
\resizebox{\linewidth}{!}{%
\begin{tabular}{lclcccc}
\toprule
Task & Episodes & Variant & T.Acc.$\uparrow$ & C.IoU$\uparrow$ & V.IoU$\uparrow$ & MAE$\downarrow$ \\
\midrule
fold clothes and bag & 9 & ET+HOI & 0.7670 & 0.4769 & 0.4395 & 0.0254 \\
 &  & No T-Rex & 0.7715 & 0.4773 & 0.4426 & 0.0246 \\
 &  & w/o distillation + data change & 0.7294 & 0.4651 & 0.4262 & 0.0248 \\
\addlinespace
move toolbox & 20 & ET+HOI & 0.7671 & 0.4608 & 0.3641 & 0.0673 \\
 &  & No T-Rex & 0.7826 & 0.4557 & 0.3592 & 0.0641 \\
 &  & w/o distillation + data change & 0.7754 & 0.4396 & 0.3450 & 0.0614 \\
\addlinespace
open close window & 5 & ET+HOI & 0.8343 & 0.5691 & 0.4794 & 0.0320 \\
 &  & No T-Rex & 0.8391 & 0.5807 & 0.4903 & 0.0315 \\
 &  & w/o distillation + data change & 0.8398 & 0.5667 & 0.4720 & 0.0318 \\
\addlinespace
pick up dairy product & 6 & ET+HOI & 0.8870 & 0.5530 & 0.4875 & 0.0363 \\
 &  & No T-Rex & 0.9056 & 0.5514 & 0.4874 & 0.0369 \\
 &  & w/o distillation + data change & 0.8794 & 0.5440 & 0.4826 & 0.0365 \\
\bottomrule
\end{tabular}
}
\end{table}

\subsection{FPHA transfer protocol and sensitivity}
\label{app:fpha_protocol}

The FPHA evaluation is independent of the earlier EgoTouch text exports. Of 287 object-annotation directories, 29 have missing or empty skeleton/object annotations; the remaining 258 sequences contain 21,501 frames with RGB, skeleton, object pose, and public MANO fits.\footnote{Public MANO fits: \texttt{fhbhands\_fits.tgz} from the handobjectconsist repository.} The main table uses only the 124 sequences and 10,998 frames belonging to the official test split. The all-available results below additionally include the official training split, evaluated without FPHA adaptation; they are not mislabeled as test results. Contact is defined on the annotated right hand by a 10 mm nearest-object-vertex distance, not by measured pressure or triangle-surface distance. The MANO fits have mean joint error 10.93 mm relative to annotated joints, so geometric labels near the contact boundary are uncertain.

The original TouchAnything checkpoint is a DINOv2 ViT-B/14 pressure-only model. It uses ImageNet RGB normalization, absolute camera-frame joint coordinates in metres, and FP32. \method{} uses its saved wrist-relative preprocessing and bfloat16; the same source RGB and joint annotations are supplied to both. Both retain the trained ego-view embedding path. The original model has no historical memory, while \method{} retains its trained cache. Inference uses non-overlapping eight-frame clips sampled at 15 Hz, repeats the final frame to complete a short clip, and discards padded outputs. Linear interpolation restores predictions to the 30 Hz source timestamps; this is offline evaluation, not strictly causal 30 Hz streaming. The directly sampled test subset contains 5,532 timestamps. This differs from the legacy TouchAnything script's dense sliding-window aggregation.

For both checkpoints, continuous normalized pressure is projected before applying the strict $>0.1$ contact threshold; it is not interpreted as calibrated contact probability. AUROC is undefined for 1,373 single-class test timestamps and is averaged over the remaining 9,625. IoU, precision, recall, and F1 assign zero to zero denominators and average over all 10,998 timestamps. Both evaluations reuse byte-identical contact labels and the same mapping file. The primary projection assigns each MANO vertex the nearest supported neutral fine-mesh vertex and averages its overlapping tactile-cell values. Only 244 of 778 vertices are within 5 mm of tactile support; nearest-support distances have median 8.84 mm and maximum 49.20 mm. Consequently, full-surface extrapolation can propagate scores beyond observed sensor regions. Table~\ref{tab:fpha_sensitivity} reports fixed alternatives, without selecting a mapping on FPHA labels. The reference-pose rows are distinguished from the RGB-only rows in Table~\ref{tab:fpha}; their different input access precludes interpreting the table as an isolated architectural comparison.

\paragraph{RGB-only contact baselines.}
DECO uses the released HRNet/context checkpoint in FP32 with its official evaluation preprocessing: the whole image is resized to $256\times256$, retained in BGR channel order, scaled by $1/255$, and normalized by the repository's ImageNet constants. The 6,890 SMPL contact scores are restricted through a fixed, topology-verified correspondence to the 778 right-hand MANO vertices; no contact-label-dependent fit is used. Contact is probability $\geq0.5$. The unnormalized demo preprocessing is a separate diagnostic and is not used in Table~\ref{tab:fpha}.

HACO uses the released HaMeR-based checkpoint in FP32 and native 778-vertex contact probabilities, with the fixed $>0.52$ threshold. The WiLoR detector receives BGR arrays according to its API convention; the subsequent hand crop uses RGB. We preserve the released right-hand selection and $2\times$ crop expansion, with a full-image fallback when detection fails, and omit demo-only temporal smoothing and connected-component filtering. No GT hand box or pose is supplied. On the native 30 Hz test grid, right-hand detection succeeds in 5,198 of 10,998 frames; all frames remain in evaluation. For both baselines, the primary table selects the same 5,532 direct test timestamps as the pressure models, interpolates continuous scores to all 10,998 frames, and applies the threshold afterwards. The label cache, AUROC exclusions, and zero-denominator conventions are shared across all four methods.

\begin{table}[!t]
\caption{\textbf{FPHA pressure-head sensitivity analysis.} Primary mapping is shown on all-available, official-test, and directly sampled subsets, followed by support-zero and covered-only test variants. Each paired comparison uses shared timestamps, labels, and thresholds.}
\label{tab:fpha_sensitivity}
\centering
\small
\setlength{\tabcolsep}{4pt}
\begin{tabular}{lccccc}
\toprule
Method & AUROC$\uparrow$ & IoU$\uparrow$ & Precision$\uparrow$ & Recall$\uparrow$ & F1$\uparrow$ \\
\midrule
\multicolumn{6}{l}{\emph{All available: 258 sequences, 21,501 frames; primary mapping}} \\
TouchAnything & .5544 & .0731 & .2035 & .1161 & .1233 \\
\method{} & .7366 & .2431 & .2959 & .6070 & .3553 \\
\midrule
\multicolumn{6}{l}{\emph{Official test: 10,998 frames; primary mapping}} \\
TouchAnything & .5486 & .0715 & .2011 & .1153 & .1209 \\
\method{} & .7431 & .2446 & .2933 & .6169 & .3558 \\
\midrule
\multicolumn{6}{l}{\emph{Test: 5,532 directly sampled timestamps; primary mapping}} \\
TouchAnything & .5483 & .0715 & .2004 & .1152 & .1209 \\
\method{} & .7416 & .2427 & .2921 & .6091 & .3533 \\
\midrule
\multicolumn{6}{l}{\emph{Test: 10,998 frames; support-zero, all 778 vertices}} \\
TouchAnything & .5319 & .0311 & .2406 & .0376 & .0570 \\
\method{} & .5504 & .1277 & .3344 & .1882 & .2124 \\
\midrule
\multicolumn{6}{l}{\emph{Test: 10,998 frames; covered-only, 244 vertices}} \\
TouchAnything & .5531 & .0682 & .2406 & .1002 & .1154 \\
\method{} & .7586 & .2554 & .3344 & .5732 & .3613 \\
\bottomrule
\end{tabular}
\end{table}

\subsection{Interpretation of ablation runs}

The five recorded configurations use the same 40-episode unseen evaluation subset with ego-only input and WiLoR pose. The no-distillation run also changes its training mixture, while the memory-on run has a different initialization from the other four rows. The evaluated best checkpoints occur at different epochs. These reports therefore summarize the available configurations but do not isolate the effects of privileged views, distillation, memory, or source diversity. The archived scalar reports contain no independent contact-head outputs or frame-level event statistics.

\section{Protocol for collected tactile recordings}
\label{app:benchmark_protocol}

\subsection{Task catalog}
\label{app:benchmark_tasks}

Table~\ref{tab:benchmark_tasks} enumerates the complete collection rather than only the tasks present in the cached evaluation export. Task descriptions specify the instructed manipulation and its main contact sequence; they are not frame-level success labels. The final column identifies tactile changes that the task was designed to expose, but does not assert that every sequence contains each event. Operation family, complexity, and expected tactile value are description-based strata: C1 denotes a basic action, C2 a contact-constrained or organized action, and C3 a compound coordinated action; L, M, and H denote lower, moderate, and high expected diagnostic value of touch, respectively. These labels organize analysis and are not empirical difficulty or tactile-dependence measurements.

\begingroup
\scriptsize
\setlength{\tabcolsep}{2.5pt}
\renewcommand{\arraystretch}{1.02}
\begin{longtable}{@{}>{\raggedright\arraybackslash}p{0.94in}r >{\raggedright\arraybackslash}p{0.68in}>{\raggedright\arraybackslash}p{1.55in}>{\raggedright\arraybackslash}p{1.55in}@{}}
\caption{\textbf{Complete task catalog for the collected benchmark.} $N$ is the number of retained sequences and sums to 897. A dagger marks the first 21 task identities represented in the cached raw 860-episode evaluation cohort; their retained counts sum to 828 because that raw cohort also contains 32 sequences later removed from the collection. The remaining nine tasks have 69 retained sequences but no cached predictions. ``Family / C / V'' gives operation family, complexity, and expected tactile-value stratum.}\label{tab:benchmark_tasks}\\
\toprule
Task & $N$ & Family / C / V & Instructed manipulation & Diagnostic tactile changes \\
\midrule
\endfirsthead
\multicolumn{5}{l}{\tablename~\thetable\ (continued)}\\
\toprule
Task & $N$ & Family / C / V & Instructed manipulation & Diagnostic tactile changes \\
\midrule
\endhead
\midrule
\multicolumn{5}{r}{Continued on next page}\\
\endfoot
\bottomrule
\endlastfoot
Move bottle$^{\dagger}$ & 53 & Grasp/move\newline C1 / M & Grasp a bottle, move it, and release it. & Grip establishment, sustained load, and redistribution before release. \\
Move box$^{\dagger}$ & 10 & Grasp/move\newline C1 / L & Grasp and relocate a cardboard box. & Broad contact establishment and load transfer. \\
Cap pen$^{\dagger}$ & 50 & Insert/fit\newline C2 / H & Align a cap with a pen and press it into place. & Misalignment resistance, local loading, and insertion termination. \\
Plug/unplug charger$^{\dagger}$ & 50 & Insert/fit\newline C2 / H & Insert a charger connector and remove it again. & Alignment, constrained insertion, seating, and separation. \\
Classify objects$^{\dagger}$ & 49 & Grasp/move\newline C2 / M & Repeatedly pick, sort, and place multiple objects. & Object-dependent grip, repeated contact transitions, and release timing. \\
Grasp parcel box$^{\dagger}$ & 50 & Grasp/move\newline C1 / L & Establish and maintain a grasp on a parcel box. & Broad contact onset and stable holding. \\
Fill pen holder$^{\dagger}$ & 50 & Insert/fit\newline C2 / M & Align a pen with a holder, lower it inside, and release it. & Rim contact, entry, and transfer from hand support to the holder. \\
Open glasses$^{\dagger}$ & 20 & Open/close\newline C2 / M & Hold the glasses and unfold their temples. & Fine grasp, hinge resistance, and finger-contact transfer. \\
Tabletop grasp$^{\dagger}$ & 50 & Grasp/move\newline C1 / L & Pick up an object from a tabletop and put it down. & Contact onset, grasp maintenance, and release. \\
Type on keyboard$^{\dagger}$ & 50 & Press/touch\newline C1 / H & Produce repeated key presses on a keyboard. & Brief fingertip contacts and repeated load--release pulses. \\
Click mouse$^{\dagger}$ & 59 & Press/touch\newline C1 / H & Hold a mouse and perform repeated clicks. & Sustained holding plus short press--release pulses and their timing. \\
Open/close notebook$^{\dagger}$ & 39 & Open/close\newline C2 / M & Open and close a hinged notebook or computer. & Hinge resistance and contact migration during reversal. \\
Take out and start laptop$^{\dagger}$ & 39 & Open/close\newline C3 / M & Take out a laptop, open it, and press its power control. & Contact transfer across carrying, hinge motion, and a brief button press. \\
Open camera stand$^{\dagger}$ & 32 & Open/close\newline C2 / M & Unfold or open a camera support. & Mechanism resistance, regrasping, and motion endpoints. \\
Open box$^{\dagger}$ & 10 & Open/close\newline C2 / M & Hold a box and separate or open its lid. & Constraint resistance followed by opening and separation. \\
Grasp phone$^{\dagger}$ & 30 & Grasp/move\newline C1 / M & Pick up and hold a mobile phone. & Contact location, grip maintenance, and redistribution. \\
Plug/unplug connector$^{\dagger}$ & 30 & Insert/fit\newline C2 / H & Align, insert, and remove a plug. & Constrained fitting, seating, load change, and detachment. \\
Press stapler$^{\dagger}$ & 48 & Press/touch\newline C2 / H & Load a stapler downward and release it. & Increasing load, peak response, mechanism transition, and unloading. \\
Two-hand laptop support$^{\dagger}$ & 30 & Grasp/move\newline C1 / M & Lift or hold a laptop with both hands. & Bilateral support, sustained contact, and load redistribution between hands. \\
Stand bottle upright$^{\dagger}$ & 50 & Grasp/move\newline C2 / M & Reorient a bottle upright, place it on the table, and let go. & Transition from hand support to environmental support and release timing. \\
Wipe table$^{\dagger}$ & 29 & Wipe\newline C2 / H & Wipe a tabletop with a paper tissue while maintaining surface contact. & Sustained contact, relative-intensity variation, and intermittent contact loss. \\
Lift cable & 6 & Grasp/move\newline C1 / M & Grasp a flexible cable and raise it from the surface. & Localized grip establishment and sustained holding. \\
Remove adapter & 6 & Insert/fit\newline C2 / H & Grasp an adapter or connector and pull it free. & Constrained extraction, changing resistance, and the instant of separation. \\
Lift package & 6 & Grasp/move\newline C1 / L & Grasp a package and raise it from the surface. & Broad contact onset and transfer of load to the hands. \\
Handover and open earbud case & 6 & Open/close\newline C3 / H & Pick up an earbud case, transfer it between hands, and open the lid. & Receiving-hand support, releasing-hand offset, fine grasp, and hinge loading. \\
Left-hand bottle lift & 8 & Grasp/move\newline C1 / M & Pick up and raise a bottle with the left hand. & Unilateral grip onset, sustained loading, and contact stability. \\
Right-hand tissue lift & 9 & Grasp/move\newline C1 / M & Pick up and raise a paper tissue with the right hand. & Light localized contact and thin-object grasp establishment. \\
Right-hand phone touch & 9 & Press/touch\newline C1 / L & Approach and touch the side of a phone with the right hand. & Near-contact versus true contact and the contact-onset boundary. \\
Right-hand keyboard press & 9 & Press/touch\newline C1 / H & Press downward on a keyboard with the right hand. & Short contact onset, relative-intensity pulse, and release. \\
Wipe glasses & 10 & Wipe\newline C3 / H & Hold glasses with one hand while wiping them with a tissue in the other. & Complementary bilateral contact, stable support, and light sliding contact. \\
\end{longtable}
\endgroup
\noindent\begin{minipage}{\linewidth}
\captionsetup{type=table,hypcap=false}
\caption{\textbf{Sensor-grounded tactile prediction by operation complexity.} Each cell compares TouchAnything / TouchAnnotator on the same 860-episode, 21-task cohort as Table~\ref{tab:bench_main}. C1 (basic), C2 (contact-constrained), and C3 (compound) are preliminary task-description labels; the first four metrics are percentages and MAE uses normalized 0--1 responses. Bold marks the better result within each stratum.}
\label{tab:bench_complexity}
\centering
\footnotesize
\setlength{\tabcolsep}{3pt}
\renewcommand{\arraystretch}{1.06}
\begin{tabular*}{\linewidth}{@{\extracolsep{\fill}}lcccc@{}}
\toprule
Metric (TA / Ours) & Overall (21/860) & C1 (8/340) & C2 (12/480) & C3 (1/40) \\
\midrule
T.Acc.$\uparrow$ & 63.29/\textbf{68.67} & 56.83/\textbf{71.99} & \textbf{68.40}/66.85 & \textbf{62.61}/62.21 \\
T.F1$\uparrow$ & 76.87/\textbf{81.08} & 72.42/\textbf{83.70} & \textbf{80.99}/79.91 & 66.14/\textbf{70.90} \\
C.IoU$\uparrow$ & 6.89/\textbf{10.73} & 6.46/\textbf{12.37} & 7.45/\textbf{9.04} & 5.51/\textbf{8.99} \\
V.IoU$\uparrow$ & 9.10/\textbf{14.12} & 8.43/\textbf{15.45} & 9.85/\textbf{12.79} & 8.36/\textbf{13.19} \\
MAE$\downarrow$ & \textbf{0.083013}/0.087488 & \textbf{0.097578}/0.103156 & \textbf{0.075393}/0.078276 & \textbf{0.051389}/0.061773 \\
\bottomrule
\end{tabular*}
\end{minipage}

\paragraph{Alignment and measured support.}
Egocentric RGB, bimanual 21-joint geometry, and tactile streams are recorded in ROS2 MCAP files. Tactile arrays are paired with sensor-header messages by MCAP log time, then aligned with camera frames by ROS header timestamps. The alignment script defaults to 2 ms header pairing and 10 ms camera--tactile tolerance. Six Wuji glove zones are baseline-corrected and projected onto the canonical $21\times21$ grid. Table~\ref{tab:bench_main} uses a 217-cell distal/proximal pressure layout per hand, with 214 cells having measurement support on a valid hand; three unsupported placeholders are excluded. Native finger rows are split by fixed 4:5 distal/proximal proportions for the long fingers and 4:2 for the thumb, without anatomical boundary calibration. The earlier 151-cell pressure projection defines a different target and is not mixed with these scores. Stored glove pose is used for alignment; the model input uses ego-estimated pose.

\paragraph{Relative-pressure scale.}
Measured responses from both hands are divided by their shared maximum over the synchronized trajectory; model predictions retain their native $[0,1]$ scale. This offline normalization uses sensor ground truth to define trajectory-relative targets, preserving within-trajectory intensity changes while removing absolute scale between trajectories. It does not establish physical pressure units or provide test-time calibration to the annotator.

\paragraph{Aggregate metrics.}
For measured and predicted values $Y,\hat Y$ on the same supported GT cells, Contact IoU compares $Y>0.1$ with $\hat Y>0.1$; Volumetric IoU is $\sum_i\min(Y_i,\hat Y_i)/\sum_i\max(Y_i,\hat Y_i)$; and MAE measures their absolute difference. Measured zeros participate, missing measurements remain excluded, and missing predictions at supported cells cause evaluation to fail. These are map metrics, separate from contact-head scores. Table~\ref{tab:bench_main} uses pooled micro aggregation: sufficient statistics are summed over all episodes in a stratum before each ratio is computed. Temporal accuracy and F1 compare frame-level contact presence: at least $5\%$ of the supported split-GT cells across available hands, including the proximal band, must exceed $0.1$. The second model band was labeled bend-associated during training, so its score against projected proximal pressure does not establish matching physical target semantics.

\paragraph{Temporal metric scope.}
Temporal accuracy and temporal F1 assess frame-level contact presence, but neither directly measures onset or offset delay. The cached aggregate exports do not contain the frame-level predictions needed to analyze transition timing or calibrate relative pressure.

\paragraph{Collection and evaluated cohort.}
The 897 retained sequences across 30 tasks are collection-list totals. Table~\ref{tab:bench_main} instead evaluates a cached raw cohort of 860 episodes from the first 21 tasks. Of these, 32 were later dropped from the collection; 69 retained episodes from nine later tasks have no cached predictions in this cohort. Thus the table describes its stated cohort and does not measure performance over all retained sequences. Participant, object, and session counts and split independence are not established by these aggregate exports.

\section{Limitations}
\label{app:limitations}

\paragraph{Limitations.}
Pressure is not uniquely observable from RGB, and sensor-relative targets do not establish force calibration across devices. Geometry-derived contact depends on reconstruction and threshold choice; canonicalization also loses sensor-specific spatial detail. Predictions depend on ego-estimated pose and may fail under blur, truncation, or prolonged occlusion. Clip-level causality permits bidirectional processing within each observed clip and does not provide frame-online output. Benefits for downstream robot policies remain untested.

\end{document}